\documentclass[letterpaper, 10 pt, conference]{ieeeconf}  
\IEEEoverridecommandlockouts                              
\usepackage[utf8]{inputenc}
\usepackage[T1]{fontenc}
\usepackage[hidelinks]{hyperref}
\usepackage{graphicx}
\usepackage{amsmath,amssymb}
\usepackage{xspace}
\usepackage[per-mode=symbol,binary-units=true,range-units=single,range-phrase=-,detect-weight=true,detect-family=true]{siunitx}
\usepackage{paralist}
\usepackage{comment}
\usepackage{url}
\usepackage{tikz}
\usetikzlibrary{fit,shapes.callouts,shapes.geometric,backgrounds,positioning,arrows.meta}

\graphicspath{{./figures/}}
\newcommand{\eg}{e.g.,\ }

\newcommand{\etal}{\xspace{}et al.\xspace}

\newcommand{\reffig}[1]{Fig.~\ref{#1}}

\newcommand{\refsec}[1]{Sec.~\ref{#1}}
\newcommand{\refeq}[1]{Eq.~\ref{#1}}

\DeclareMathOperator{\sort}{sort}

\title{\LARGE \bf
Trajectory Generation with Fast Lidar-based\\3D Collision Avoidance for Agile MAVs
}

\author{Marius Beul, and Sven Behnke 
\thanks{This work has been supported by the German Federal Ministry of Education and Research (BMBF) in the project ”Kompetenzzentrum: Aufbau des Deutschen Rettungsrobotik-Zentrums (A-DRZ)``}%
\thanks{Institute for Computer Science VI, Autonomous Intelligent Systems, University of Bonn, Endenicher Allee 19a, 53115 Bonn, Germany,
		{\tt\small mbeul@ais.uni-bonn.de}%
}
}

\begin{document}

\maketitle
\thispagestyle{empty}
\pagestyle{empty}

\begin{tikzpicture}[overlay, remember picture]
  \path (current page.north) ++(0.0,-1.0) node[draw = black] {Accepted for IEEE International Symposium on Safety, Security, and Rescue Robotics (SSRR), Abu Dhabi, UAE, 2020};
\end{tikzpicture}
\vspace{-0.3cm}

\begin{abstract}
Micro aerial vehicles (MAVs), are frequently used for exploration, examination, and surveillance during search and rescue missions. Manually piloting these robots under stressful conditions provokes pilot errors and can result in crashes with disastrous consequences. Also, during fully autonomous flight, planned high-level trajectories can be erroneous and steer the robot into obstacles.

In this work, we propose an approach to efficiently compute smooth, time-optimal trajectories MAVs that avoid obstacles. Our method first computes a trajectory from the start to an arbitrary target state, including position, velocity, and acceleration. It respects input- and state-constraints and is thus dynamically feasible. Afterward, we efficiently check the trajectory for collisions in the 3D-point cloud, recorded with the onboard lidar. We exploit the piecewise polynomial formulation of our trajectories to analytically compute axis-aligned bounding boxes (AABB) to speed up the collision checking process. If collisions occur, we generate a set of alternative trajectories in real-time. Alternative trajectories bring the MAV in a safe state, while still pursuing the original goal. Subsequently, we choose and execute the best collision-free alternative trajectory based on a distance metric.

The evaluation in simulation and during a real firefighting exercise shows the capability of our method.
\end{abstract}

\section{Introduction}
\label{sec:Introduction}
Micro aerial vehicles (MAVs) are becoming a key element in reducing the required risks, time, and costs for search and rescue missions, aerial reconnaissance, and disaster examination. In most cases, a human pilot operates the MAV remotely to fulfill a mission, or the MAV is following a predefined path of GPS waypoints at an altitude assumed to be obstacle-free.
However, permanent line-of-sight from the pilot to the MAV may not be maintainable at all times due to large obstacles. Also, thin obstacles like antennas or power lines may only be perceivable inaccurately by the pilot. Other (probably moving) participants in the airspace like drones, rescue helicopters, birds, or debris proposes a risk for the MAV. Furthermore, during a real rescue scenario, the operator's cognitive load is significant \cite{kruijff2014ar}, which provokes human errors.
The loss of an MAV due to a crash is expensive, but even worse, it can have disastrous consequences for the executed mission.

To tackle these challenges, in this article, we present an efficient method to compute smooth, time-optimal trajectories for MAVs that automatically avoid obstacles. On the one hand, our method can be used as an obstacle avoidance and control layer in a hierarchy of planners for fully autonomous flight. On the other hand, since it accepts intuitive setpoints, it can be employed to execute the pilot's commands during manual flight, providing an additional safety layer.
This article extends our previous work on trajectory generation for MAVs \cite{beul2017icuas} with an obstacle avoidance feature,

Our main contributions are
\begin {compactitem}
  \item computation of dynamically feasible collision-free alternative trajectories, 
  \item fast analytical bounding box computation for cropping of large point clouds,
  \item analytical modeling and integration of sensor coverage during planning, and
  \item evaluation in simulation and a real-world firefighting exercise.
\end{compactitem}

\begin{figure}[t]
  \centering
  \includegraphics[trim=00mm 00mm 00mm 00mm,clip,width=1.0\linewidth]{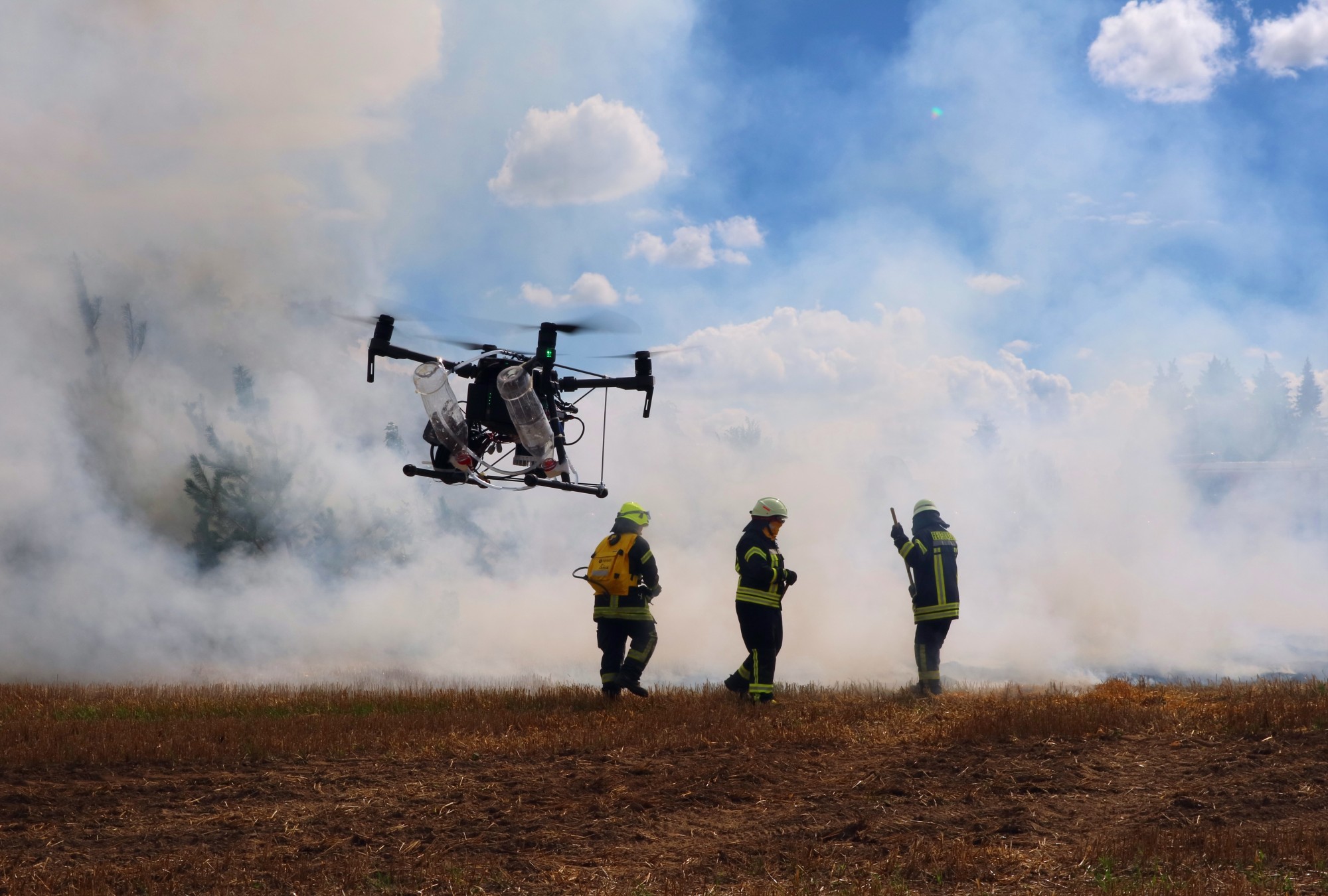}~
  \vspace{-1.0ex}
  \caption{Operation of our trajectory generation method onboard an MAV during a real firefighting exercise.}
  \label{fig:teaser}
  \vspace{-3.5ex}
\end{figure}

\reffig{fig:teaser} shows our MAV supporting firefighters by surveying the area with its multimodal sensor setup.

\section{Related Work}
\label{sec:Related_Work}
Low-level obstacle avoidance is an active field of research.

For example, Zhang \etal \cite{zhang2018iros} present a method to instantaneously avoid static obstacles by following precomputed paths in cluttered environments. The paths are hierarchically organized, such that branches from alternative paths can be efficiently stored and executed. The method is fast ($\le \SI{50}{\milli\second}$), but generated trajectories are not optimal.

Similarly, also Barry \etal \cite{barry2018jfr} show impressive results with fast flights of up to \SI{14}{\meter\per\second}. Like the above-noted approach, the authors precompute trajectories from which they select a collision-free instance during runtime. In contrast, our approach quickly generates trajectories during runtime, incorporating the current continuous 3D state of the robot.

Also, the idea to instantaneously react to sensor measurements instead of planning a trajectory is not new, and many different approaches from vector field histograms \cite{ulrich1998icra} to potential field-based obstacle avoidance have been presented.

For example, Falanga \etal \cite{falanga2020scirob} use an event camera to quickly detect obstacles in the vicinity of the MAV. While in contrast to normal RGB-cameras, event cameras are advantageous in terms of latency and dynamic range, they still suffer from a narrow field of view and low range compared to lidars. To guarantee a low latency, instead of sophisticated trajectory generation techniques, the MAV performs potential field-based obstacle avoidance.

Similarly, also Nieuwenhuisen \etal \cite{nieuwenhuisen2013isprs} perform potential field-based obstacle avoidance with an MAV. To account for the MAV's non-neglectable dynamics, the authors extend the method with a motion model and predict the MAV state in the near future. The authors report that trajectories with active motion model are smoother than with pure reactive potential field obstacle avoidance.

The idea to generate a set of alternative dynamically feasible trajectories and subsequently selecting a suitable one has been shown in, \eg \cite{mueller2015tro}. Instead of obstacle avoidance, the authors use the motion primitives to catch a flying ball with an MAV. Like our approach, the authors use closed-form solutions to find suitable trajectories, resulting in computation times in the order of microseconds. The alternative trajectories are not time-optimal and not guaranteed to be collision-free.

\begin{figure}[t]
  \centering
  \resizebox{1.0\linewidth}{!}{\input{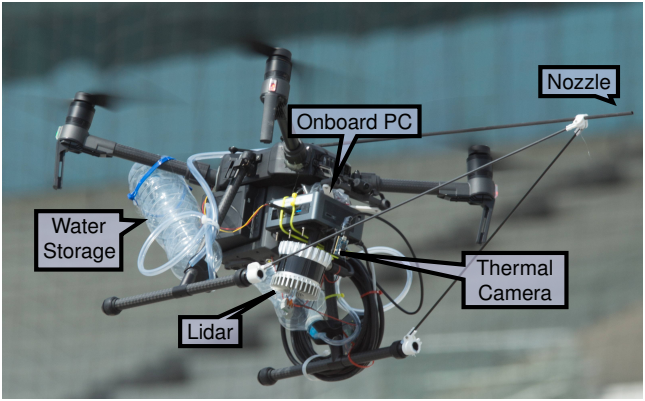}}
  \vspace{-2.0ex}
  \caption{Design of our MAV equipped with an Ouster OS1-64 Gen~1, a FLIR Lepton thermal camera, a fire extinguisher, and a lightweight but powerful onboard computer.}
  \label{fig:mav}
  \vspace{-3.5ex}
\end{figure}

Lindqvist \etal \cite{lindqvist2020ral} present an MPC that features collision avoidance and can also deal with moving obstacles. The authors use a nonlinear solver to find collision-free trajectories within the \SI{50}{\milli\second} replanning time. In contrast to our technique, the pipeline only predicts \SI{2}{\second} of the trajectory, making it unsuitable for fast flight or fast obstacles. Instead of onboard sensors, the technique relies on a known obstacle path, measured with a motion capture system.

Trajectories produced by all methods mentioned above are dynamically feasible, but \emph{not} time-optimal. To avoid nimble obstacles (or during fast flight), the control inputs of the robot have to be maximized to achieve maximum dexterity. Thus, similar to our approach, the method from Lopez and How \cite{lopez2017icra} generates time-optimal state and input constrained trajectories by sampling terminal states. With a computation time of \SIrange{2}{5}{\micro\second}, the method is real-time capable. The authors use an RGB-D camera to measure a point cloud in front of the MAV. Without omnidirectional perception (\eg obtained with a lidar), safe trajectories are forced to lie in front of the MAV. An additional drawback of the method is that it works in (constant) velocity space instead of position space; thus, \eg it restricts the MAV from hovering. The method assumes obstacles to be static.

Watterson and Kumar \cite{watterson2015iros} report a hybrid approach for obstacle avoidance. On the one hand, the proposed method uses a receding horizon control policy (RHCP) to steer an MAV through a cluttered environment. On the other hand, the approach always guarantees that there exists a safe stopping policy that brings the MAV to hover. Similarly, our approach searches for multiple classes of trajectories.

To our knowledge, no method exists that can compute smooth, time-optimal obstacle-avoiding 3D trajectories for MAVs within typical control-loop frequencies. The method proposed in this work replaces our reactive low-level obstacle avoidance mechanism \cite{beul2018iros} that does not scale to aggressive high-speed trajectories.

\section{System Setup}
\label{sec:System_Setup}
In the following sections, we first describe the hardware of our MAV in \refsec{sec:Hardware_Design}. We continue by presenting our approach to point cloud filtering with axis-aligned bounding boxes in \refsec{sec:Analytical_Bounding_Box}. Collision checking is described in \refsec{sec:Collision_Checking} and \refsec{sec:Analytical_Coverage_Test}. Lastly, we present our method for generating alternative trajectories in \refsec{sec:Alternative_Trajectories}.

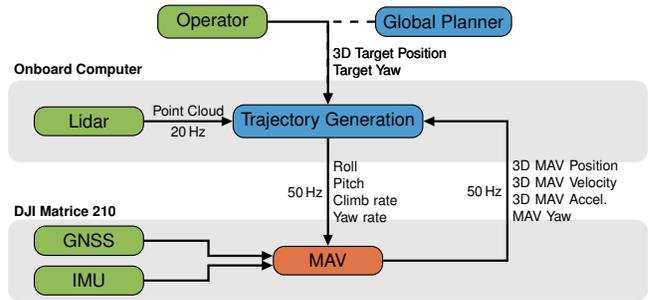
\begin{figure}[t]
  \centering
  \resizebox{1.0\linewidth}{!}{\begin{tikzpicture}[font=\sffamily,>={Stealth[inset=0pt,length=4pt,angle'=45]}]
\tikzset{content_node/.append style={minimum size=1.5em,minimum width=6em,draw,align=center,rounded corners,scale=0.65,inner sep=1.5mm}}
\tikzset{label_node/.append style={scale=0.5}}
\tikzset{group_node/.append style={dotted,align=center,rounded corners,inner sep=1em,thick}}

\definecolor{green}{rgb} {0.466, 0.674, 0.188}
\definecolor{blue}{rgb}  {0.000, 0.447, 0.741}
\definecolor{red}{rgb}   {0.850, 0.325, 0.098}
\definecolor{grey}{rgb}  {0.5,0.5,0.5}

\draw[thick, rounded corners, grey!20!white,fill] (-4.0,0.5) -- (4.0,0.5) -- (4.0,1.5) -- (-4.0,1.5) -- cycle;
\draw[thick, rounded corners, grey!20!white,fill] (-4.0,-1.25) -- (4.0,-1.25) -- (4.0,-0.25) -- (-4.0,-0.25) -- cycle;

\node(Lidar)[content_node,fill=green!80!white] at (-3.0,1.0) {Lidar};
\node(Trajectory_Generation)[content_node,fill=blue!70!white] at (0.0,1.0) {Trajectory Generation};
\node(Operator)[content_node,fill=green!80!white] at (-1.45,2.25) {Operator};
\node(Global_Planner)[content_node,fill=blue!70!white] at (1.45,2.25) {Global Planner};

\node(MAV)[content_node,fill=red!80!white] at (0.0,-1.25+0.5) {MAV};
\node(GNSS)[content_node,fill=green!80!white] at (-3.0,-1.0+0.5) {GNSS};
\node(IMU)[content_node,fill=green!80!white] at (-3.0,-1.5+0.5) {IMU};

\draw[->, thick] (Trajectory_Generation) -- node[label_node,midway,left] {50\,Hz} node[label_node,midway,right,text width=1cm] {Roll Pitch Climb~rate Yaw~rate} (MAV);
\draw[->, thick] (MAV) -- (2.25,-1.25+0.5) -- node[label_node,midway,left] {50\,Hz} node[label_node,midway,right,text width=2.8cm] {3D~MAV~Position 3D~MAV~Velocity 3D~MAV~Accel. MAV~Yaw} (2.25,1.0) -- (Trajectory_Generation);

\draw[->, thick] (GNSS) -- (GNSS -| -1.5,-1.0)  -- (-1.5,-1.0 |- MAV.175) -- node[label_node,midway,left] {} node[label_node,midway,right] {} (MAV.175);
\draw[->, thick] (IMU) --  (IMU -| -1.5,-1.0) --  (-1.5,-1.0 |- MAV.185)  -- node[label_node,midway,left] {} node[label_node,midway,right] {} (MAV.185);
\draw[->, thick] (Operator) -- (Operator -| 0.0,2.25) -- node[label_node,midway,right,align=left] {3D~Target~Position\\Target~Yaw} (Trajectory_Generation);
\draw[->, thick,dashed] (Global_Planner) -- (Global_Planner -| 0.0,2.25) -- node[label_node,midway,right,align=left] {3D~Target~Position\\Target~Yaw} (Trajectory_Generation);
\draw[->, thick] (Lidar) -- node[label_node,midway,below] {20\,Hz} node[label_node,midway,above,align=left] {Point Cloud} (Trajectory_Generation);

\node(ROS_Group_Label)[label_node,anchor=south west] at (-4.0,1.5) {\textbf{Onboard Computer}};
\node(MAV_Group_Label)[label_node,anchor=south west] at (-4.0,-0.25) {\textbf{DJI Matrice 210}};

\end{tikzpicture}}
  \vspace{-2.0ex}
  \caption{Structure of our method. Green boxes represent external inputs like sensors, blue boxes represent software modules, and the red box indicates the MAV flight control. All software components use ROS as middleware. Position, velocity, acceleration, and yaw are allocentric. Commands for our trajectory generator can either directly come from an operator or from a high-level planner.}
  \label{fig:structure}
  \vspace{-3.5ex}
\end{figure}

\subsection{Hardware Design}
\label{sec:Hardware_Design}
To support firefighters during their mission, we developed the MAV shown in \reffig{fig:mav}. It is based on the DJI Matrice~210 platform and features a lidar to perform simultaneous localization and mapping (SLAM), and a thermal camera to detect and map, \eg fires or victims. Furthermore, it features small but fast Intel Bean Canyon NUC8i7BEH onboard PC with an Intel\textsuperscript{\textregistered} Core\textsuperscript{\texttrademark} i7-8559U processor and \SI{32}{\giga\byte} of RAM. To extinguish small fires, the MAV is equipped with a fire extinguisher. We use the robot operating system (ROS) as middleware on the MAV.

\reffig{fig:structure} shows an excerpt of our software pipeline, running onboard the MAV. In the shown example, an operator directly defines 4D setpoints (X, Y, Z, Yaw) for the trajectory generator. Instead, the commands can also be generated by a higher layer planning pipeline during a fully autonomous flight (see \cite{beul2018iros}).

\begin{figure}[t]
  \centering
  \includegraphics[trim=00mm 00mm 00mm 00mm,clip,width=1.0\linewidth]{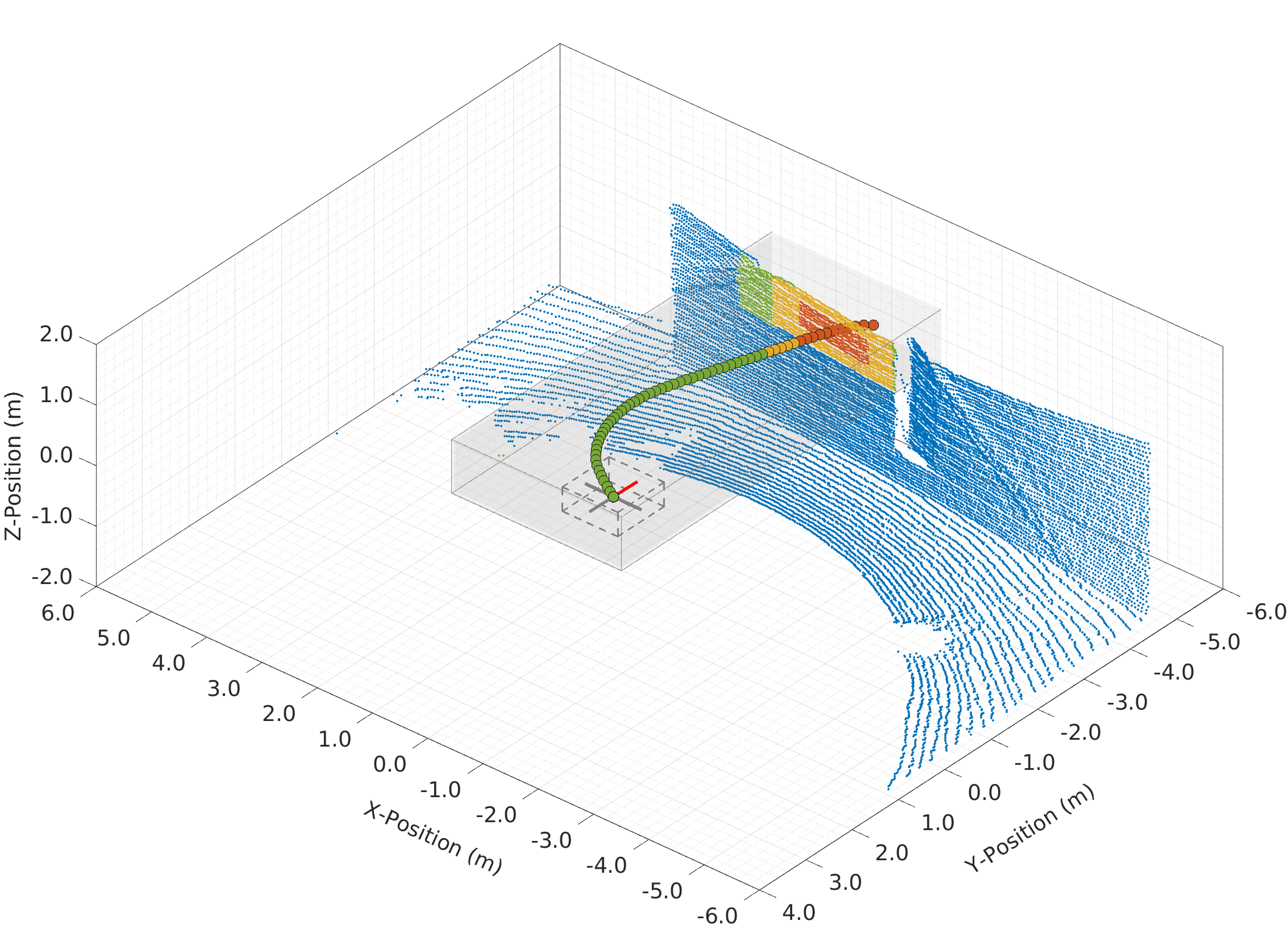}~
  \vspace{-1.0ex}
  \caption{Analytical computation of an axis-aligned bounding box (AABB) for a polynomial trajectory (gray, expanded by the MAV dimensions). The AABB is used to select a subset (green, yellow, red) of the point cloud (blue). Afterward, we sample the trajectory with a constant position $\Delta p$ and check for collisions with the MAV (red) and collisions with the MAV expanded by a warning distance (yellow). The points in the point cloud subset are safe (green), within warning distance of at least one trajectory sample (yellow), and in collision distance of at least one trajectory sample (red).}
  \label{fig:bounding_box}
  \vspace{-3.5ex}
\end{figure}

The depicted trajectory generator\footnote{\url{https://github.com/AIS-Bonn/opt\_control}} generates smooth third-order time-optimal trajectories that respect asymmetrical state and input constraints. It is described in detail in \cite{beul2017icuas}, \cite{beul2016icuas}, and \cite{beul2019iros}. We now extend the capabilities of our trajectory generator by real-time obstacle avoidance employing point clouds recorded by the onboard lidar.

Since our lidar produces up to \num{65536} measurements per scan (\num{1310720} per second), we first demonstrate how we analytically compute bounding boxes to crop the large point clouds in \refsec{sec:Analytical_Bounding_Box}. We then describe our measurement model and how we check if trajectories pass unknown space in \refsec{sec:Analytical_Coverage_Test}. In \refsec{sec:Constant_Distance_Sampling}, we show how we exploit the polynomial formulation of our trajectories to analytically sample the trajectories with a constant position offset (per dimension). Subsequently, we present how we generate alternative collision-free trajectories that bring the MAV in a safe state while still pursuing the original goal in \refsec{sec:Alternative_Trajectories}.

\subsection{Analytical Bounding Box}
\label{sec:Analytical_Bounding_Box}

\begin{figure}[t]
  \centering
  \resizebox{0.8\linewidth}{!}{\rotatebox{10}{\begin{tikzpicture}[font=\sffamily,>={Stealth[inset=0pt,length=10pt,angle'=45]}]
\tikzset{content_node/.append style={minimum size=1.5em,minimum width=6em,draw,align=center,rounded corners,scale=0.65,inner sep=1.5mm}}
\tikzset{label_node/.append style={scale=0.5}}
\tikzset{group_node/.append style={dotted,align=center,rounded corners,inner sep=1em,thick}}

\definecolor{green}{rgb} {0.466, 0.674, 0.188}
\definecolor{blue}{rgb}  {0.000, 0.447, 0.741}
\definecolor{red}{rgb}   {0.850, 0.325, 0.098}
\definecolor{grey}{rgb}  {0.5,0.5,0.5}


\draw[line width=1.0pt,fill=red!80!white,fill opacity=0.8] (0.0,0.0) -- (-{atan(5/9)}+180:6) arc(-{atan(5/9)}+180:{atan(5/9)}:6) -- cycle;

\draw[line width=1.0pt] (0.0,0.0) -- (-{atan(5/9)}:-2) arc(-{atan(5/9)}:{atan(5/9)}:-2) -- cycle;
\node[anchor=south] at (-1.2,-0.2) {$\theta_{lidar}$};

\draw[line width=2.0pt,color=black,dashed] (5.0,2.0) -- (0.0,2.0);

\draw[->,line width=2.0pt,color=blue!70!white] (0.0,1.6) -- (3.65,1.6);
\node[anchor=west] at (3.65,1.6) {$l_{cone}$};
\draw[->,line width=2.0pt,color=green!80!white] (0.0,1.2) -- (5.0,1.2);
\node[anchor=west] at (5.0,1.2) {$l_{test}$};

\draw[->, line width=2.0pt,color=grey!20!white] (0.0,0.0) -- (0.0,1.0);
\node[anchor=east] at (0.0,0.8) {$p_{norm}$};

\draw[->, line width=2.0pt,color=grey!20!white,dashed] (0.0,0.0) -- (0.0,6.0);
\node[anchor=east] at (0.0,3.0) {$l_{lidar}$};

\draw[fill=black](5.0,2.0)circle(3.0pt);
\node[anchor=south] at (5.0,2.0) {$p_{test}$};

\draw[fill=black](0.0,2.0)circle(3.0pt);
\node[anchor=south east] at (0.0,2.0) {$p_{is}$};

\draw[fill=black](3.65,2.0)circle(3.0pt);

\draw[->,line width=2.0pt,color=black] (0.0,0.0) -- ({5.0*cos(10)},-{5.0*sin(10)});
\node[anchor=north,rotate=-10] at ({5.0*cos(10)},-{5.0*sin(10)}) {X};
\draw[->,line width=2.0pt,color=black] (0.0,0.0) -- ({6.5*sin(10)},{6.5*cos(10)});
\node[anchor=east,rotate=-10] at ({6.5*sin(10)},{6.5*cos(10)}) {Y};

\end{tikzpicture}}}
  \vspace{-4.0ex}
  \caption{Simplified schematic of our lidar model projected onto a 2D-plane. By checking if $l_{test} < l_{cone}$, we can determine if the point $p_{test}$ is inside unobservable space. For simplicity, we only show one of two cones.}
  \label{fig:cone_schematic}
  \vspace{-3.5ex}
\end{figure}

To speed up computation, we first filter the point cloud. Instead of removing random points or other filtering methods, we crop the point cloud such that points that are outside of an axis-aligned bounding box (AABB) around the trajectory are discarded.
To do so, we analytically compute the axis-aligned bounding box employing the polynomial formulation of the trajectory.

As described in \cite{beul2017icuas}, our trajectories consist of a concatenation of n-dimensional polynomials with constant n-dimensional jerk. In each segment and each dimension, the position $p$ follows \refeq{eq:analytical_bounding_box_1} with the jerk $j$, the acceleration $a$, and the velocity $v$ at the beginning of the segment. We determine the global maxima and minima of each polynomial by deriving the position with respect to time. With $t_{ex}$ being the segment's extremal times from \refeq{eq:analytical_bounding_box_2} and \refeq{eq:analytical_bounding_box_3}, we subsequently check if the extremal time lies within the segment's interval. If so, the segment produces a global extremum of the trajectory at point $p_{ex}$ with \refeq{eq:analytical_bounding_box_4}. We execute this procedure for each dimension for each segment to compute the axis-aligned bounding box. As a last step, we expand the bounding box by the MAV size.
\begin{align}
  p(t) &= v \, t + \tfrac{1}{2} \, a \, t^2 + \tfrac{1}{6} \, j \, t^3                      \label{eq:analytical_bounding_box_1}\\
  p'(t_{ex}) &:= 0 \Rightarrow \nonumber \\
  0 &= v + a \, t_{ex} + \tfrac{1}{2} \, j \, t_{ex}^2                                      \label{eq:analytical_bounding_box_2}\\
  t_{ex} &= -\frac{(a \pm \sqrt{(a^2 - 2 \, j \, v)}}{j}                                    \label{eq:analytical_bounding_box_3}\\
  p_{ex} &= p + v \, t_{ex} + \tfrac{1}{2} \, a \, t_{ex}^2 + \tfrac{1}{6} \, j \, t_{ex}^3 \label{eq:analytical_bounding_box_4}
\end{align}
\reffig{fig:bounding_box} displays an example AABB. In this typical example, only \num{1372} of the \num{65536} total scan points are in the proximity of the trajectory and considered for the following computations. This means a reduction of \SI{97.9}{\percent}.

\begin{figure}[t]
  \centering
  \includegraphics[trim=00mm 00mm 00mm 00mm,clip,width=1.0\linewidth]{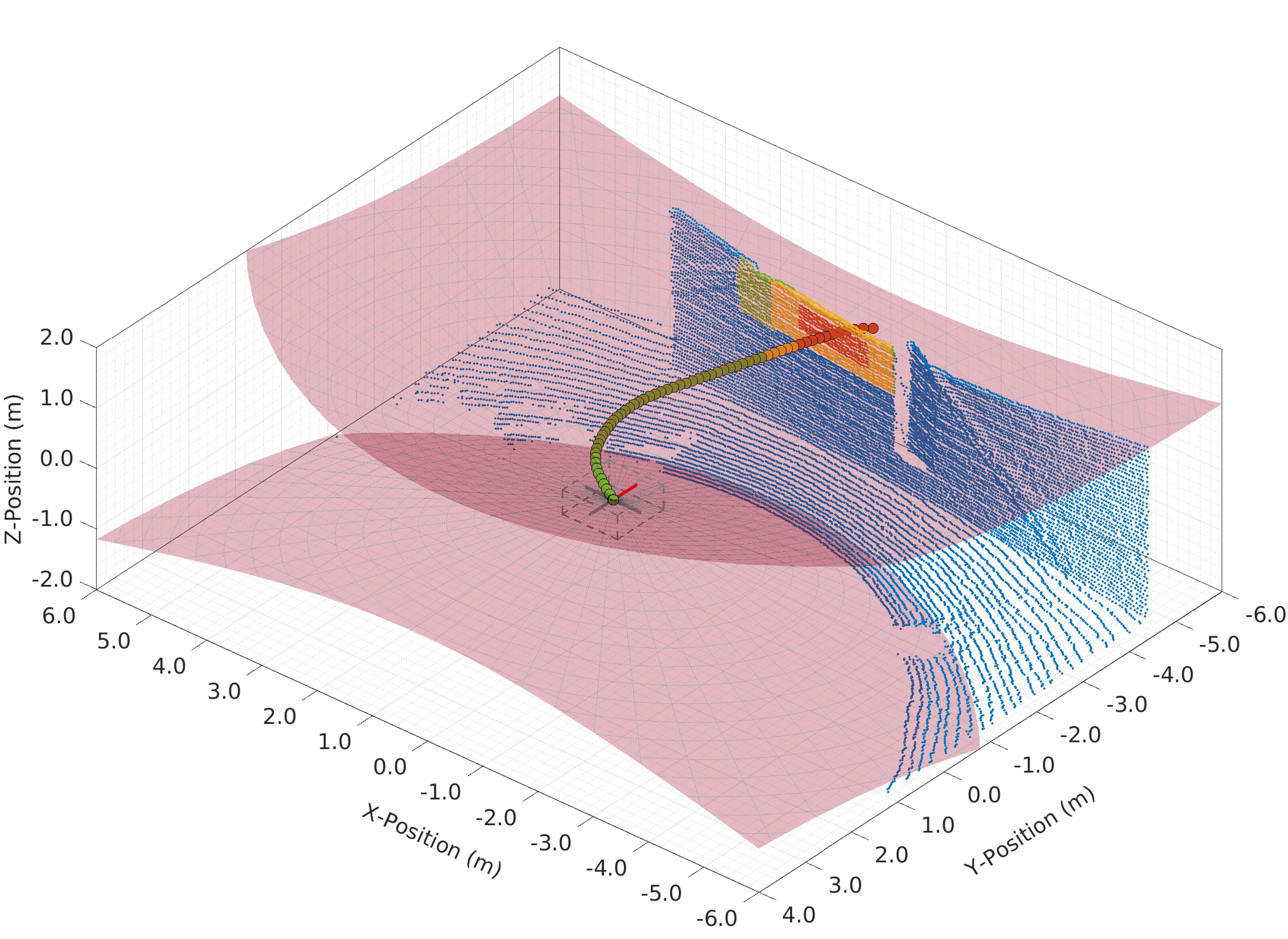}~
  \vspace{-1.0ex}
  \caption{Analytical lidar coverage computation. We model the lidar sensor coverage by two cones (red) with a \SI{33.2}{\degree} opening angle. The range of the lidar is \SI{120}{\meter}. We detect and reject trajectories that pass through unobservable space.}
  \label{fig:cone}
  \vspace{-3.5ex}
\end{figure}

\subsection{Constant Distance Sampling}
\label{sec:Constant_Distance_Sampling}
We now sample the trajectory with a constant time difference, or with a constant position difference. We depict sampling with a constant time difference in \refeq{eq:constant_time_sampling}, with $p_{n+1}$ being the position after $\Delta t$ with the initial position $p_{n}$, velocity $v_{n}$, acceleration $a_{n}$, and jerk $j_{n}$.
\begin{align}
  p_{n+1} &= p_{n} +  v_{n} \, \Delta t + \tfrac{1}{2} \, a_{n} \, \Delta t^2 + \tfrac{1}{6} \, j_{n} \, \Delta t^3 \label{eq:constant_time_sampling}
\end{align}

However, sampling the trajectory with a constant time difference gives inhomogeneous resolution over the trajectory. For collision checking, we thus prefer the (more complicated) constant distance sampling. To do so, we analytically solve \refeq{eq:constant_distance_sampling} for $t$ in a preprocessing step.
\begin{align}
  \Delta p &= v_{n} \, t + \tfrac{1}{2} \, a_{n} \, t^2 + \tfrac{1}{6} \, j_{n} \, t^3 \label{eq:constant_distance_sampling}
\end{align}
This gives three analytical equations for $t_{1-3}$. For brevity, we refrain from picturing the solutions here. Subsequently, during runtime, we evaluate the three solutions with the desired position offset $\Delta p$, and select the smallest positive non-imaginary $t$ of all dimensions. We compute the position at this timestep with \refeq{eq:constant_time_sampling} as the next sample and continue this procedure until the end of the trajectory is reached. Exemplarily, \reffig{fig:bounding_box} shows a trajectory sampled with $\Delta p = \SI{10}{\centi\meter}$ in each dimension. Since trajectories can have multiple jerk segments (\eg the trajectory in \reffig{fig:bounding_box} has 18 segments), we also account for effects at the segment borders that we can not cover here for brevity.
The distance offset in each dimension does not have to be equal. Thus by assigning inhomogeneous $\Delta p$, individual dimensions can be sampled denser or sparser.

\subsection{Collision Checking}
\label{sec:Collision_Checking}
With the trimmed point cloud from \refsec{sec:Analytical_Bounding_Box}, and the trajectory positions equidistantly sampled from \refsec{sec:Constant_Distance_Sampling}, we check each dimension $n$ of each trajectory position $p_{test,n}$ for two distances $l_{test} = l_{warn} \lor l_{coll}$ indicating if a point $p_{pc,n}$ is within warning distance or if a point is so close that it causes a collision.
Checking for two individual distances allows for hysteresis at the border of obstacles and for a more sophisticated validation (see \refsec{sec:Alternative_Trajectories}).
\begin{align}
  \begin{split}
    b_{close,n} =\,&p_{pc,n} > p_{test,n} - l_{test,n} \quad \land \label{eq:collision_checking_0}\\
                   &p_{pc,n} < p_{test,n} + l_{test,n}
  \end{split}
\end{align}
Since the world is not static, we model moving observations with a constant velocity model. Although our lidar does not give 3D velocities for the measurements, future sensor modalities like radar possibly will do so, and thus we extend the approach to collision checking to points moving with the 3D velocity vector $v_{pc,n}$.

\begin{figure}[t]
  \centering
  \includegraphics[trim=00mm 00mm 00mm 00mm,clip,width=1.0\linewidth]{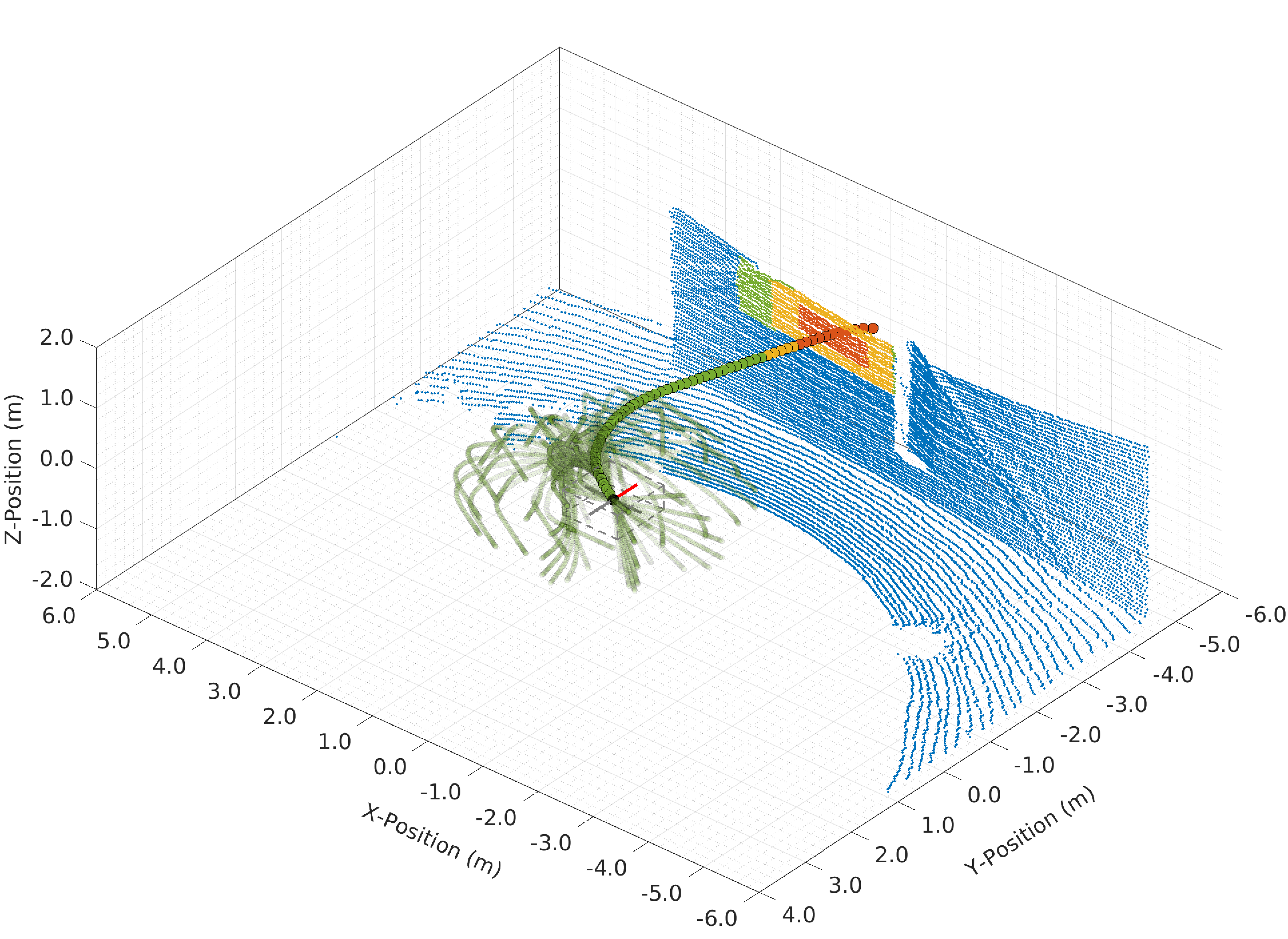}~
  \vspace{-1.0ex}
  \caption{Alternative safe trajectories that end in concentric spheroids around the MAV. All 67 trajectories are safe in terms of collisions with the point cloud \emph{and} do not pass unobservable space.}
  \label{fig:alternative_sphere}
  \vspace{-3.5ex}
\end{figure}

For each dimension $n$, we calculate the times $t_{1,n}$ and $t_{2,n}$, when the point cloud point with position $p_{pc,n}$ and velocity $v_{pc,n}$ enters resp., leaves the axis-aligned hyperrectangle defined by center point $p_{test,n}$ and elongation $l_{test,n}$ with \refeq{eq:collision_checking_1} and \refeq{eq:collision_checking_2}. Subsequently, we sort all times according to their size and check if there exists a period where all dimensions of the line are inside the hyperrectangle, as shown in \refeq{eq:collision_checking_3} and \refeq{eq:collision_checking_4}. If this is the case, the point cloud point moves through the AABB. Thus, if the line defined by $p_{pc}$ and $v_{pc}$ crosses the hyperrectangle defined by $p_{test}$ and $l_{test}$, it is causing a collision with $b_{close} = \text{true}$.
\begin{align}
  t_{1,n} &= \frac{p_{test,n}-l_{test,n}-p_{pc,n}}{v_{pc,n}} \label{eq:collision_checking_1}\\
  t_{2,n} &= \frac{p_{test,n}+l_{test,n}-p_{pc,n}}{v_{pc,n}} \label{eq:collision_checking_2}\\
  t_{sort} &= \sort(t_{n})                                   \label{eq:collision_checking_3}\\
  b_{close} &= \max(t_{sort,1}) < \min(t_{sort,2})           \label{eq:collision_checking_4}
\end{align}
In this paper, we mark each trajectory sample within a warning distance of at least one point (and the corresponding points) yellow. Points that are causing a collision, and the corresponding trajectory points, are marked red.

\subsection{Analytical Coverage Test}
\label{sec:Analytical_Coverage_Test}
Besides checking for collisions with the (potentially moving) point cloud, we also check if the trajectory passes unobservable terrain. For this, we model the 3D lidar with an omnidirectional field of view that has cone-shaped blind spots on top and bottom depicted in \reffig{fig:cone_schematic}. We also model the maximum range of the lidar. To test if a 3D point lies within the blind cones, we first obtain the normal vector of the MAV $p_{norm}$ from the Inertial Measurement Unit (IMU) which points upward during hover.
\begin{align}
  p_{is} &= (p_{norm} \boldsymbol{\cdot} p_{test}) \cdot p_{norm}          \label{eq:coverage_1}\\                                                                                   
  l_{cone} &= |p_{is}| \cdot \tan(\frac{\pi}{2}-\frac{\theta_{lidar}}{2}) \label{eq:coverage_2}\\
  l_{test} &= |p_{test} - p_{is}|                                          \label{eq:coverage_3}\\
  l_{test} &< l_{cone} \land p_{is,z} > 0 \Rightarrow \text{In upper cone} \label{eq:coverage_4}\\
  l_{test} &< l_{cone} \land p_{is,z} < 0 \Rightarrow \text{In lower cone} \label{eq:coverage_5}\\
  l_{test} &> l_{lidar} \qquad\qquad\;\;\:\Rightarrow \text{Out of range}  \label{eq:coverage_6}
\end{align}
We then compute the point on the normal vector $p_{is}$ that is closest to the tested trajectory point $p_{test}$, as shown in \refeq{eq:coverage_1}. With the lidar opening angle $\theta_{lidar}$, we obtain the horizontal extend of the cone at the specific height of $p_{is}$ with \refeq{eq:coverage_2}. Subsequently, we decide if point $p_{test}$ is inside the unobservable area with \refeq{eq:coverage_3}--\refeq{eq:coverage_6}.

Since both cones touch at the origin, the model is very susceptible for small z-values. Therefore, we assume that all points that lie inside of the MAV are always collision-free. The trajectory from \reffig{fig:cone} conforms to the highlighted constraints with opening angle $\theta_{lidar} = \SI{33.2}{\degree}$ and maximum range $l_{lidar} = \SI{120}{\meter}$.

\begin{figure}[t]
  \centering
  \includegraphics[trim=00mm 00mm 00mm 00mm,clip,width=1.0\linewidth]{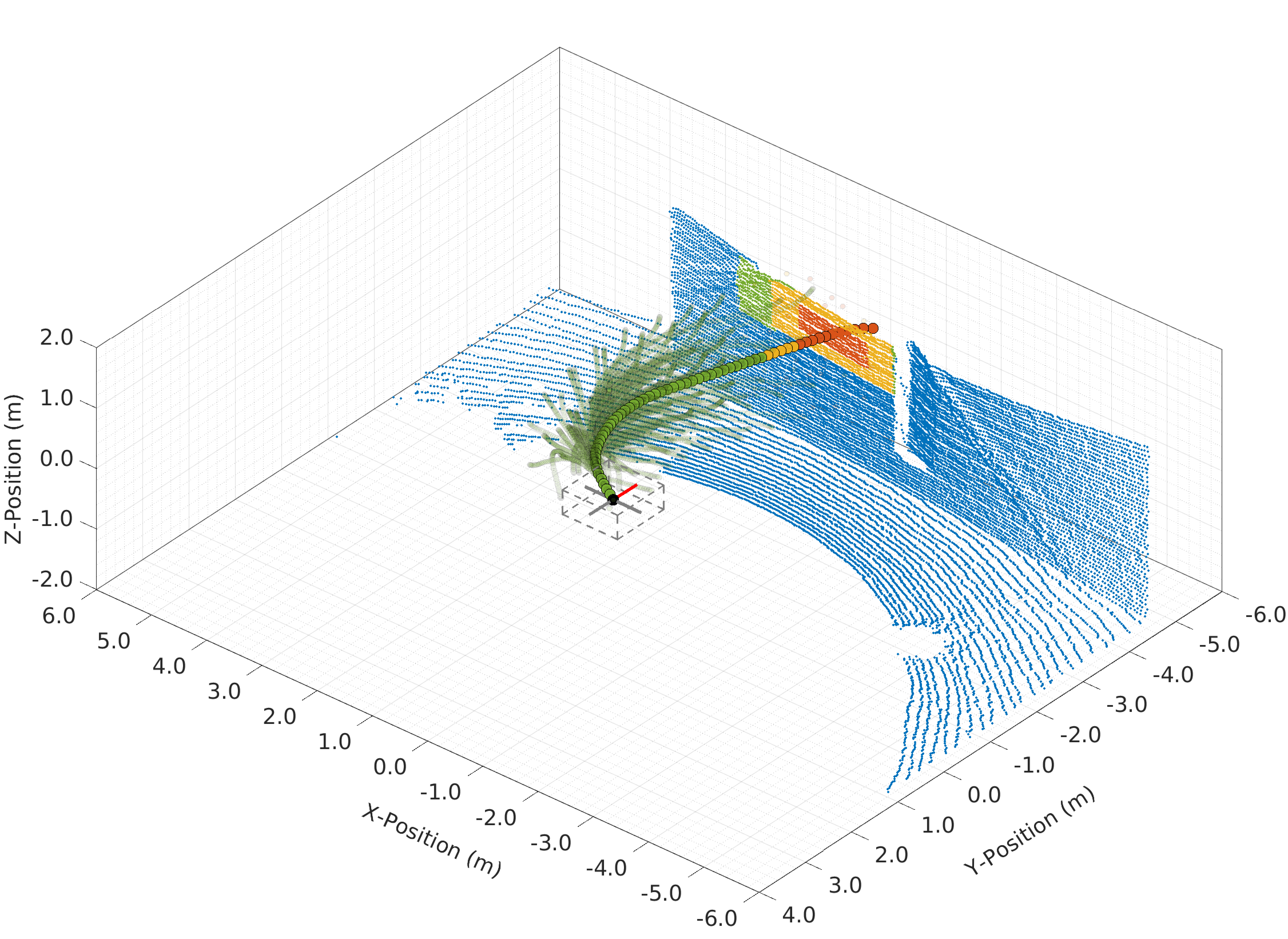}~
  \vspace{-1.0ex}
  \caption{Alternative safe trajectories that end in a concentric tube around the original trajectory. All 204 trajectories are safe in terms of collisions with the point cloud \emph{and} do not pass unobservable space.}
  \label{fig:alternative_circle}
  \vspace{-3.5ex}
\end{figure}

\subsection{Alternative Trajectories}
\label{sec:Alternative_Trajectories}
With a confirmed collision somewhere on the trajectory or a warning that will \emph{not} clear itself on the current trajectory, we generate alternative safe trajectories. If the trajectory is within warning distance at the beginning, but enters a safe state on its own, we classify it as safe. Since our trajectory generation method is very fast ($\ll \SI{1}{\milli\second}$), we can generate multiple alternative trajectories with the lidar frequency of \SI{20}{\hertz}. One can imagine many different metrics that would yield good target points for safe trajectories. We pursue two strategies. Our safe trajectories target a) waypoints on concentric spheroids around the current MAV position, and b) waypoints on concentric tubes around the original trajectory.

\reffig{fig:alternative_sphere} shows a set of alternative trajectories targeting waypoints that lie on concentric spheroids. These time-optimal trajectories bring the MAV to stop as fast as possible without bringing it significantly nearer to the original target.

Alternatively, trajectories targeting a concentric tube around the original trajectory can be safe, while still pursuing the intend of the original trajectory. \reffig{fig:alternative_circle} shows an example of these trajectories.

After generating all alternative trajectories, we execute the steps from \refsec{sec:Analytical_Bounding_Box}--\refsec{sec:Analytical_Coverage_Test} to check if each trajectory is safe. 
As with the original trajectory, an alternative trajectory is safe if it does not enter a warning (or even collision) zone around any point and does not enter unperceivable space. Leaving a warning zone, however, is permitted, since we want to be able to recover from states where the MAV is already inside a warning zone. From the set of safe alternative trajectories, we select and execute the one with the closest Euclidean distance to the original waypoint.

The number of alternative trajectory candidates determines the total computation time needed for the approach. Thus, we either fix the number of trajectory candidates considered in each replanning step, or adaptively generate alternative trajectories until the replanning time (of in our case \SI{50}{\milli\second}) is reached.

\begin{figure}[t]
  \centering
  \includegraphics[trim=00mm 00mm 00mm 00mm,clip,width=1.0\linewidth]{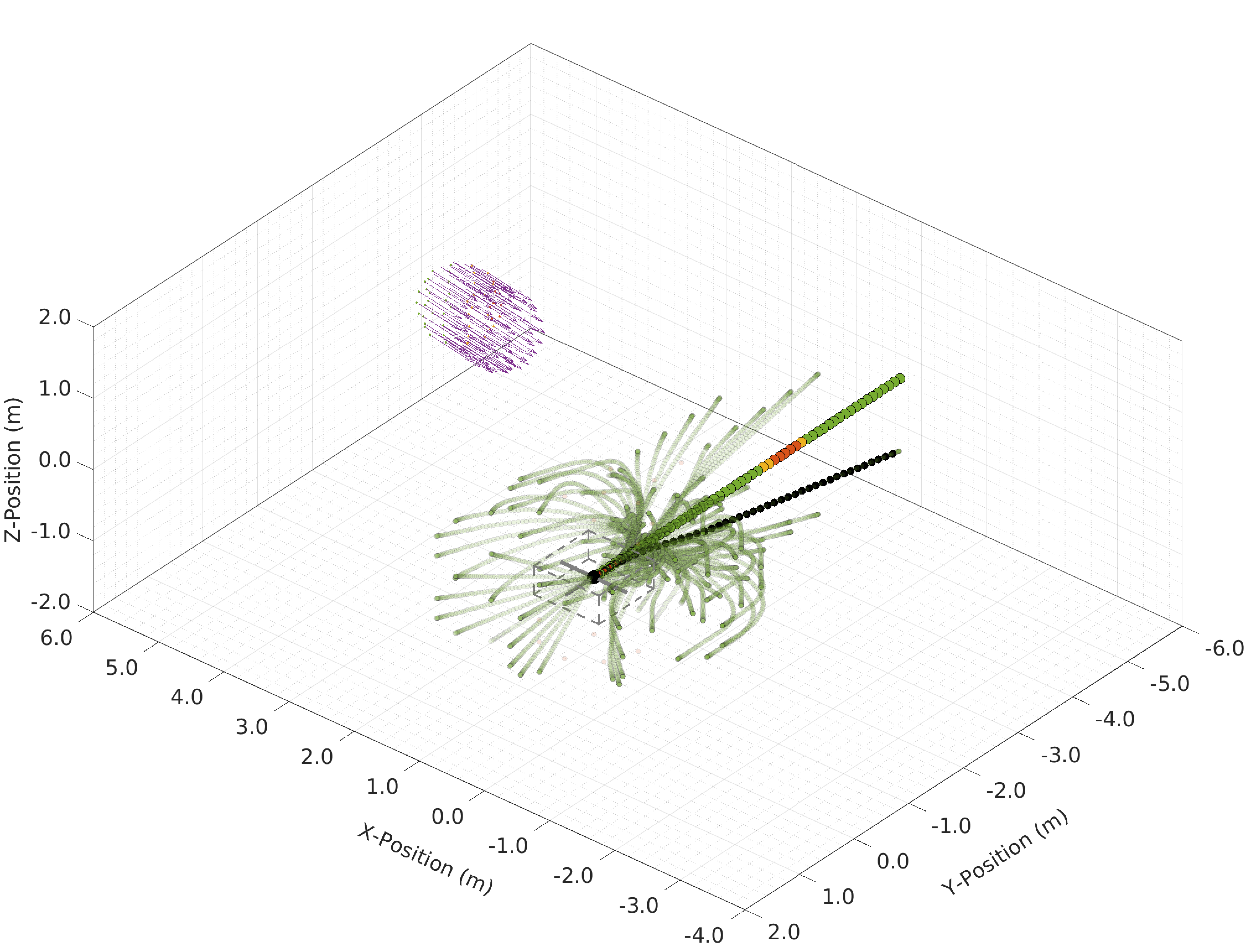}~
  \vspace{-1.0ex}
  \caption{Avoidance of a simulated obstacle moving with \SI{1.25}{\meter\per\second}. Every point in the point cloud can have an associated velocity. Moving points are predicted forward, so that collisions with the original trajectory as well as potential rescue trajectories can be predicted. Here, the MAV avoids the obstacle thrown into its original trajectory. Our method considers 44 potential rescue trajectories from which the black one is executed.}
  \label{fig:ball}
  \vspace{-3.5ex}
\end{figure}

\section{Evaluation}
\label{sec:Evaluation}
\begin{figure*}[t]
  \centering
  \includegraphics[trim=00mm 00mm 00mm 00mm,clip,width=1.0\linewidth]{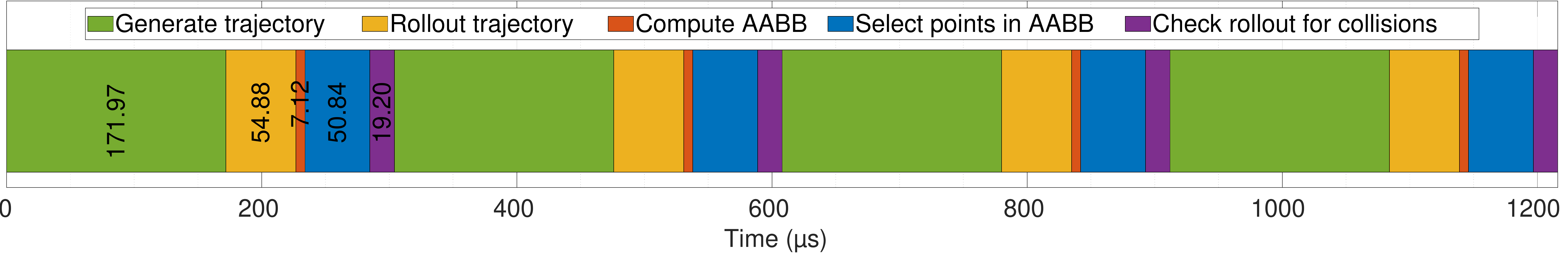}~
  \vspace{-2ex}
  \caption{Approximate computation time of our approach for a scenario with three alternative trajectories. Most time is spent on computing the individual trajectories (green). Due to the computation of the AABB (yellow), collision checking for each rollout (violet) is relatively fast. The 5-step process can be repeated arbitrarily, with approximately \SI{304}{\micro\second} per iteration.}
  \label{fig:computation_time}
  \vspace{-3ex}
\end{figure*}

We evaluate our approach in simulation as well as with a real MAV. To evaluate our approach, we use point clouds recorded during a fully autonomous flight during the Mohamed Bin Zayed International Robotics Challenge (MBZIRC) 2020. Here, our MAV flew into a wall and crashed due to a bug in the coordinate system alignment. During the challenge, we employed the same trajectory generator we present here, but unfortunately, without any obstacle avoidance. We report results from the challenge in \cite{schwarz2020jfr} and \cite{beul2020ssrr}. We previously showed the applicability of our method to MAV flight in, \eg \cite{beul2018iros}.
The laser scans from the scenarios in \reffig{fig:bounding_box}, \ref{fig:cone}, \ref{fig:alternative_sphere}, and \ref{fig:alternative_circle} are extracted from recordings of the challenge. A video showcasing the evaluation can be found on our website\footnote{\url{www.ais.uni-bonn.de/videos/ssrr_2020_beul}}.

\subsection{Computation Time}
\label{sec:Computation_Time}
We evaluate the computation time with the computer onboard the MAV. \reffig{fig:computation_time} shows the total computation time of our approach for a typical scenario.

It can be seen that most of the time is spent on computing the time-optimal trajectories and that we are able to generate, rollout, and check a total of approximately \num{164} trajectories within the \SI{50}{\milli\second} time frame.

To evaluate the effectiveness of subsetting the point cloud with the AABB, we conducted an ablation study. We removed this specific feature and directly checked for collisions of each rolled out state with the entire point cloud. By comparing the computation times for the typical scenario from \reffig{fig:computation_time}, we found that the process takes averagely \SI{1147}{\micro\second} instead of the \SI{77.16}{\micro\second} ($\SI{7.12}{\micro\second} + \SI{50.84}{\micro\second} + \SI{19.20}{\micro\second}$) from \reffig{fig:computation_time}. This corresponds to a speedup of \SI{1487}{\percent} for the collision checking method and \SI{452}{\percent} speedup in relation to the entire algorithm.

To further speed up our pipeline, we plan to adopt the approach from \cite{bialkowski2016ijrr} for collision checking with moving obstacles. Furthermore, the structure of our method offers excellent possibilities for parallelization. Thus, we are currently working on parallelizing our technique with alternative waypoints distributed to individual CPU cores.

\subsection{Moving Obstacles}
\label{sec:Moving_Obstacles}
As described in \refsec{sec:Collision_Checking}, every point in the point cloud can have an associated 3D velocity. To evaluate our approach, we simulate lidar measurements on an object that moves into the trajectory. We show the scenario in \reffig{fig:ball}. It can be seen that the obstacle crosses the trajectory. Instantaneously, the MAV generates multiple valid rescue trajectories and avoids the obstacle. It decides that it is safe and dynamically feasible to fly in front of the obstacle in the target direction (using the tube trajectories). If the obstacle moved faster, trajectories behind the obstacle would be feasible. If the object were more extensive or additional obstacles would be present, a hard braking maneuver that uses the full dynamic capabilities of the MAV would be performed.

\subsection{Real-World Experiments}
\label{sec:Real_World_Experiments}
\begin{figure}[t]
  \centering
  \includegraphics[trim=50mm 95mm 150mm 75mm,clip,width=1.0\linewidth]{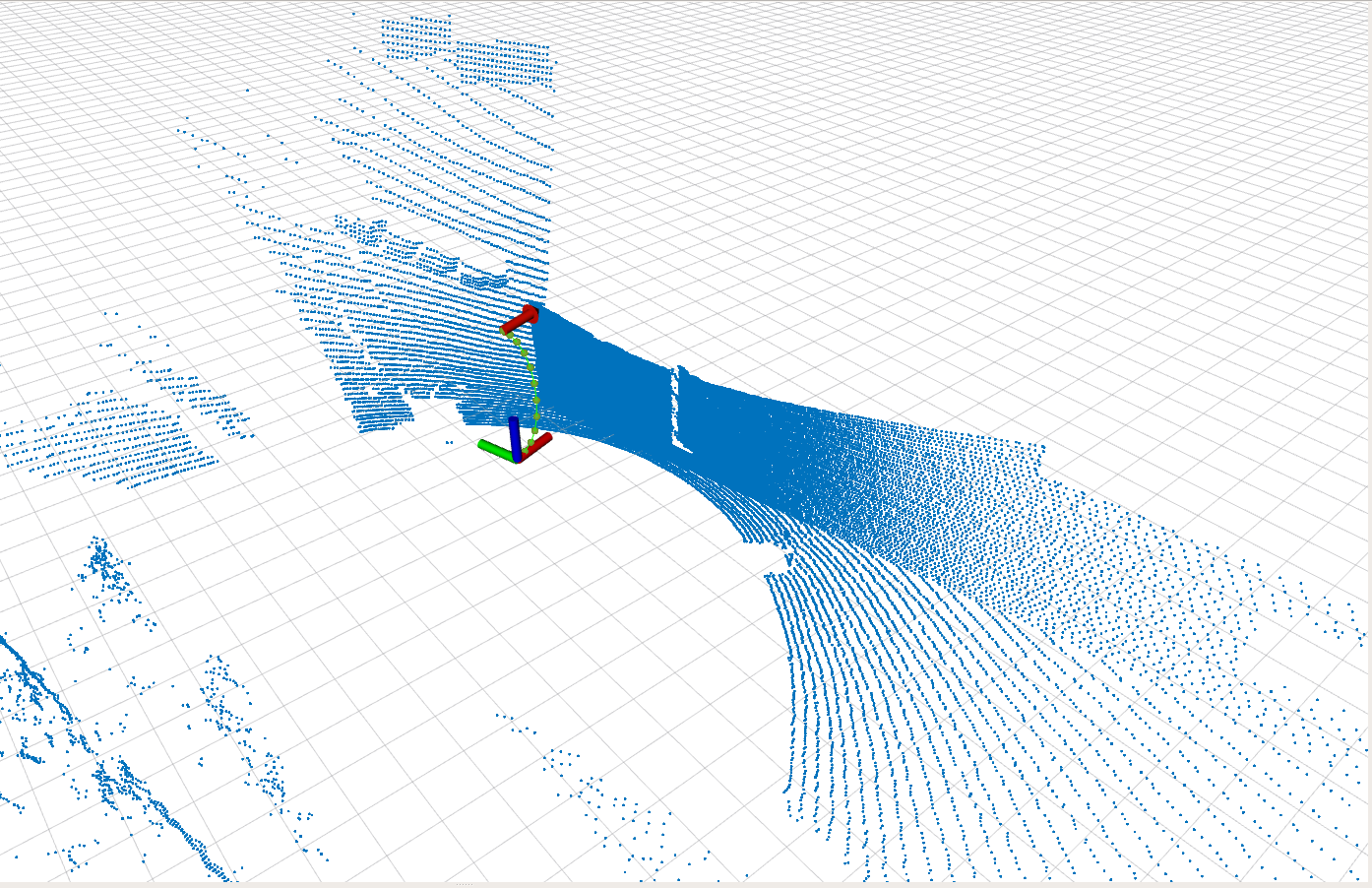}\\
  \vspace{1mm}
  \includegraphics[trim=50mm 95mm 150mm 75mm,clip,width=1.0\linewidth]{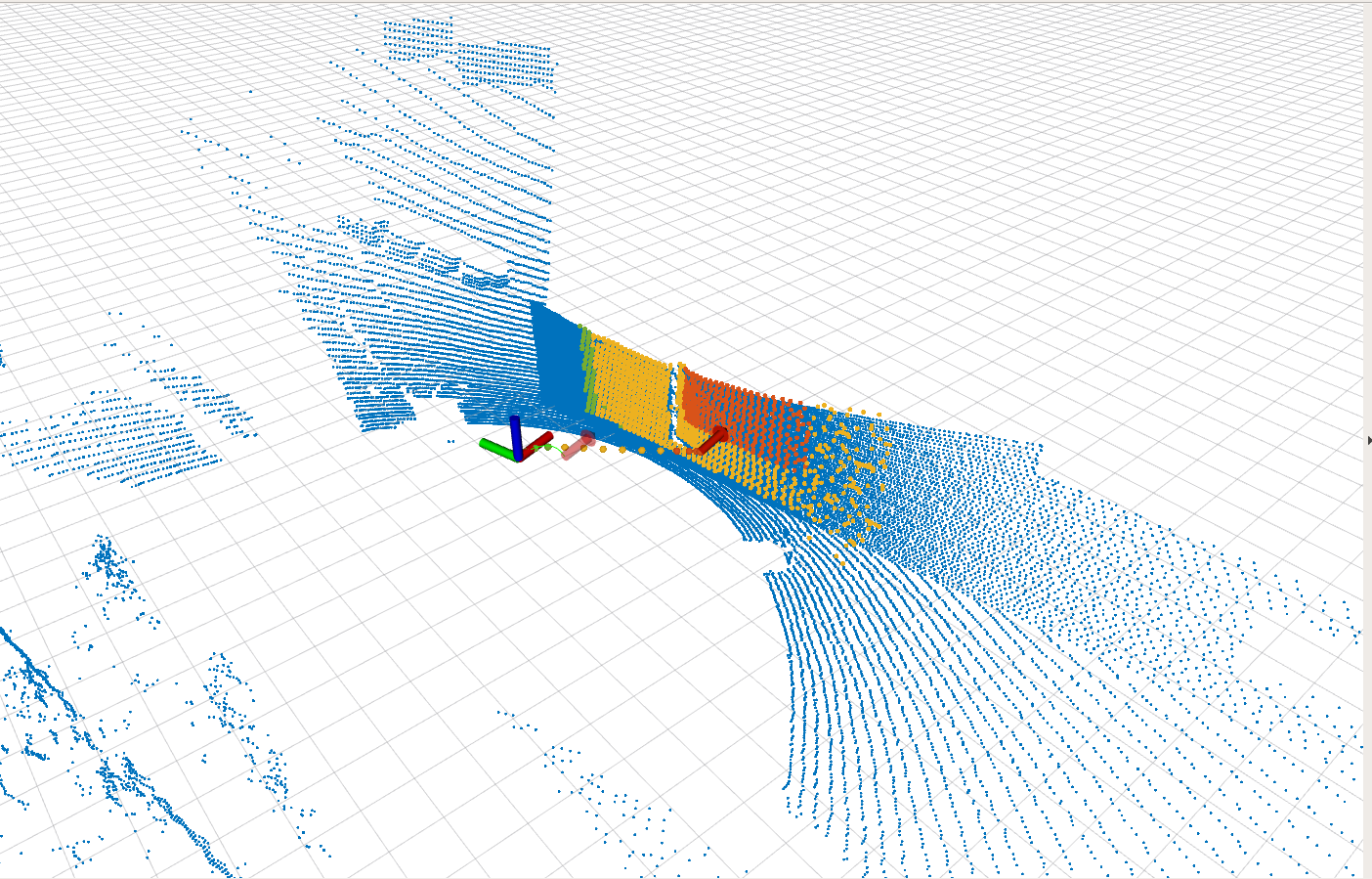}~
  \vspace{-1.0ex}
  \caption{Evaluation with the MBZIRC 2020 dataset. Top: The waypoint (red arrow) can be reached by the MAV (axis) with the green trajectory without colliding with the measured point cloud (blue). Bottom: If the waypoint lies close to the perceived wall, an alternative waypoint (transparent red) is computed that can be reached without interfering.}
  \label{fig:eval_mbzirc}
  \vspace{-3.5ex}
\end{figure}

We evaluate our method with a dataset that was recorded during the MBZIRC 2020 challenge, where our MAV actually flew into a wall and crashed. We use the recorded lidar and GPS measurements as input to our method and assess if the crash could be hindered by our new obstacle avoidance feature. \reffig{fig:eval_mbzirc} shows a still from the world model during the challenge. It can be seen that the MAV successfully avoids a collision by targeting an alternative instead of the commanded waypoint.

The experiments with the dataset show that our method never plans trajectories that lie close to measured obstacles. Instead, it always generates feasible alternative waypoints and corresponding trajectories. This experiment is, however, not sufficient to show that our method works reliably, since it does not close the loop and thus does not execute the planned trajectories.

\begin{figure}[t]
  \centering
  \includegraphics[trim=00mm 40mm 00mm 00mm,clip,width=1.0\linewidth]{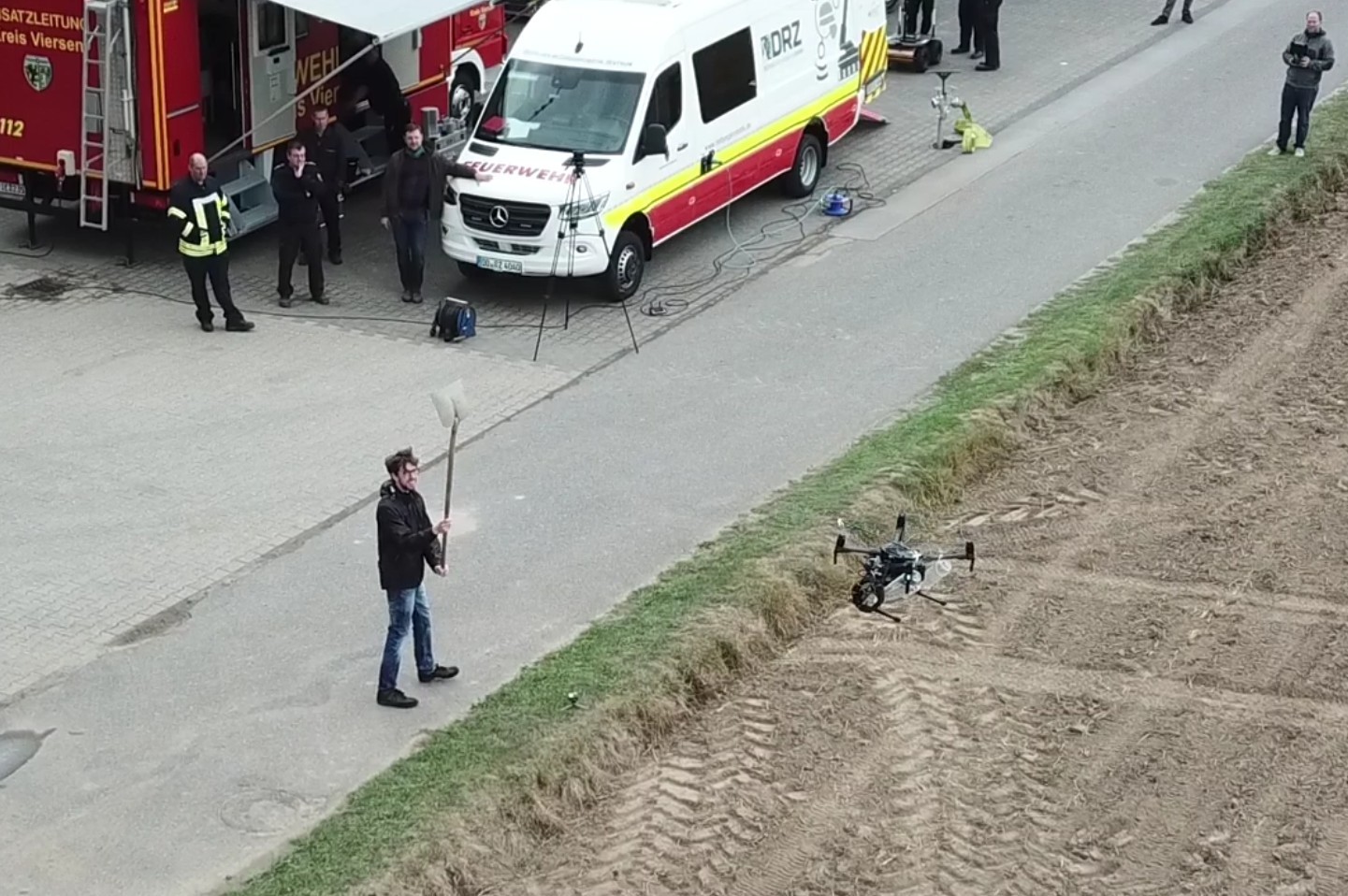}\\
  \vspace{1mm}
  \includegraphics[trim=00mm 00mm 00mm 10mm,clip,width=1.0\linewidth]{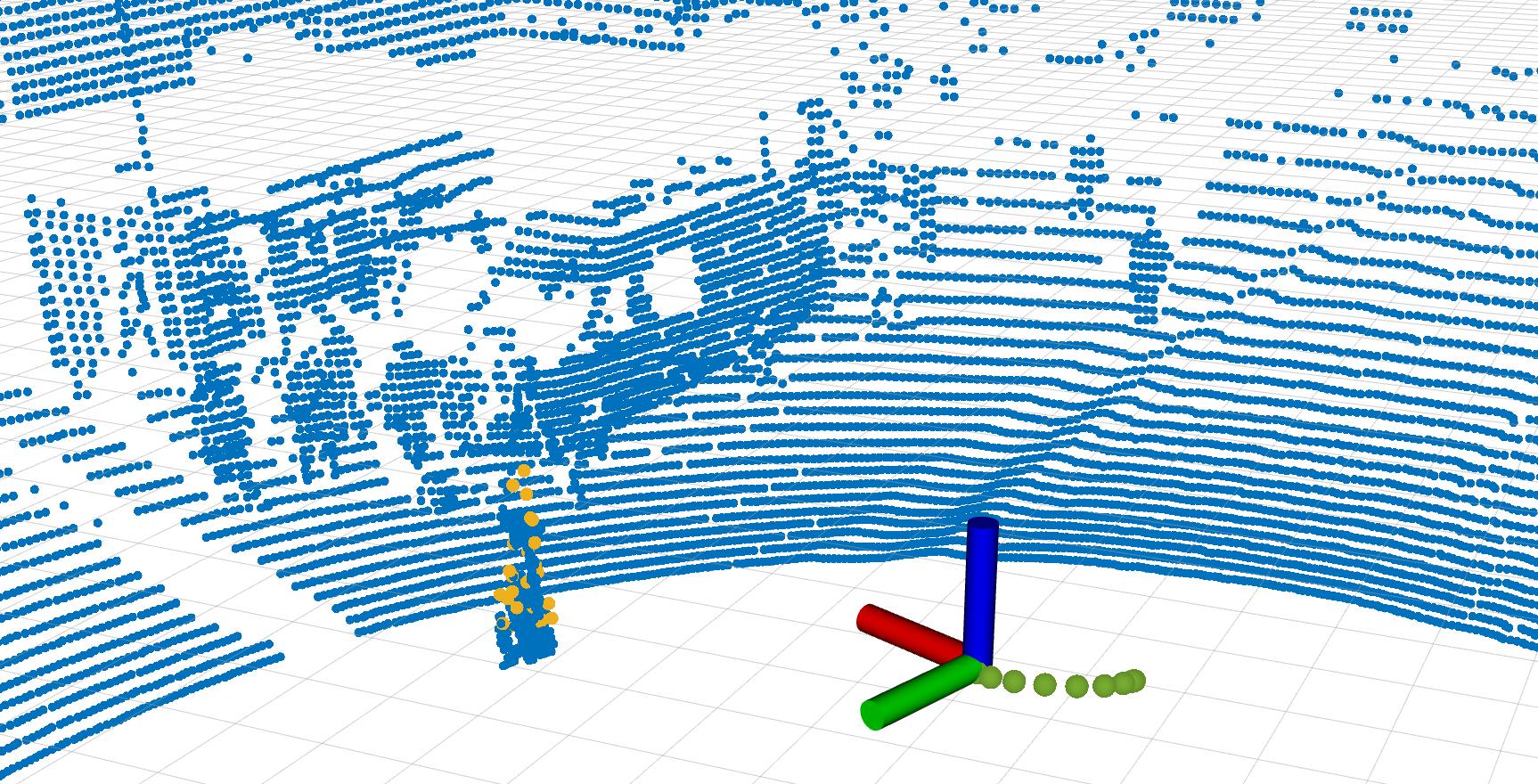}~
  \vspace{-1.0ex}
  \caption{Avoidance of an approaching person. Our MAV (axis) senses the threat (yellow points) and plans an avoidance trajectory (green).}
  \label{fig:eval_viersen}
  \vspace{-3.5ex}
\end{figure}

To evaluate our method in a real-world scenario in a closed-loop, we brought our MAV to a real firefighting exercise. Here, we a) intentionally commanded the MAV to fly into a tree, and b) approached the MAV during hovering to provoke an evasive maneuver. \reffig{fig:eval_viersen} shows one of the experiments and the corresponding world model. The MAV detects the approaching person and avoids it with a trajectory in the opposing direction. During the entire exercise, no crash with static nor dynamic obstacles happened. The closed-loop avoidance maneuvers, however, were not always smooth since the MAV often switched between alternative waypoints, causing a jittery behavior. We address this issue by favoring alternative waypoints that lie close to previously selected alternative waypoints for the alternative waypoint costs $C_{awp} = \alpha \cdot \lvert AC \rvert + (1-\alpha) \cdot \lvert AB \rvert$ with $\lvert AC \rvert$ beeing the distance of the alternative waypoint to the commanded waypoint and $\lvert AB \rvert$ beeing the distance from the alternative waypoint to the previously selected alternative waypoint. By computing more alternative trajectories in parallel, we want to increase the waypoint density in the future and thus further reduce jumps between individual alternatives.
We also found that the horizontal and vertical dynamics of our MAV differ significantly. Therefore, we had to adjust the parameters of the spheroids during the experiments to be oblate (height $\ll$ diameter) since the MAV strongly preferred vertical movement.

\section{Conclusion}
\label{sec:Conclusion}
We have provided detailed insight into our robust trajectory generation framework. The viability of our approach has been demonstrated in simulation as well as in a real-world scenario during a real firefighting exercise.

\par

In particular, the ability to analytically crop large point clouds, makes our method scale to larger and denser point clouds in the future. Due to the fast runtime, our method can run in real-time as MPC onboard the MAV, and it can react to obstacles, perceived with the lidar rate of \SI{20}{\hertz}. We showed that the ability to reject unskilled control inputs could help prevent crashes and thus increase the reliability of flying robots. 

\par

We believe that our contribution will make the operation of MAVs during rescue missions safer and more reliable.

\bibliographystyle{IEEEtranBST/IEEEtran}
\bibliography{literature_references}

\end{document}